\pdfoutput=1

\documentclass[11pt]{article}

\usepackage{EMNLP2023}

\usepackage{times}
\usepackage{latexsym}
\usepackage{booktabs}

\usepackage{amsmath}

\usepackage{graphicx}
\usepackage{svg}
\usepackage[T1]{fontenc}

\usepackage[utf8]{inputenc}

\usepackage{microtype}
\usepackage{inconsolata}

\usepackage[table,xcdraw]{}

\usepackage{subcaption}

\usepackage[normalem]{ulem}
\useunder{\uline}{\ul}{}

\usepackage{xurl}

\title{Test Suites Task: Evaluation of Gender Fairness in MT \\ with MuST-SHE and INES}

\author{Beatrice Savoldi, Marco Gaido, Matteo Negri, Luisa Bentivogli \\
         Fondazione Bruno Kessler \\ 
         \texttt{\{bsavoldi,mgaido,negri,bentivo\}@fbk.eu}}

\begin{document}
\maketitle
\begin{abstract}

As part of the WMT-2023 ``Test suites'' shared task, in this paper we summarize the results of two test suites evaluations: MuST-SHE$^{WMT23}$ and INES. 
By focusing on the en-de and de-en language pairs, we rely on these newly created test suites to investigate systems' ability to translate feminine and masculine gender and produce gender-inclusive translations. Furthermore we discuss metrics associated with our test suites and validate them by means of human evaluations. 
Our results indicate that systems
achieve reasonable and comparable performance in correctly translating both feminine and masculine gender forms for naturalistic gender phenomena. 
Instead, the
generation of inclusive language forms in translation
emerges as a challenging task for all the evaluated MT models, indicating room for future improvements and research on the topic.  




We make MuST-SHE$^{WMT23}$ and INES freely available, respectively at:

\url{https://mt.fbk.eu/must-she/} 

 \url{https://mt.fbk.eu/ines/}

\end{abstract}

\section{Introduction}
\label{sec:introduction}

As Machine Translation (MT) has made strides in generic performance,  there is an increasing recognition of the need to scrutinize finer, more nuanced aspects that defy assessment through traditional metrics 
computed on
generic test sets. It is within this context that the WMT Test Suites shared task emerges, aiming to provide a dedicated evaluation 
framework
to delve into specific dimensions of MT output with a laser focus. In particular, 
those representing
well-known challenges within the current MT landscape. 

% ...


In light of the above, our contribution is dedicated to the critical themes of gender bias and inclusivity in translation \citep{savoldi2021gender}.
Given the large-scale deployment of MT, such aspects
are not only relevant from a technical perspective, where gender-related errors negatively impact the accuracy of 
automatic translation. Rather, biased and non-inclusive systems can 
pose the concrete risk of under/misrepresenting gender minorities
by over-producing masculine forms, while reinforcing binary gendered expectations and stereotypes \citep{blodgett-etal-2020-language,lardelli-gromann-2022-gender-fair}. 

Accordingly, in this paper we present the FBK  participation in the Test 
Suites
shared task by conducting evaluations 
on two newly-created test suites: 
\begin{enumerate}
    \item 
    \textbf{MuST-SHE}$^{WMT23}$
    for en-de, created as a English$\rightarrow$German extension of the already existing multilingual MuST-SHE corpus \citep{bentivogli-2020-gender}. This dataset is designed to allow for fine-grained analysis of (binary) gender bias in MT.  

    \item \textbf{INES} for de-en, designed to assess the ability of MT systems to generate inclusive language forms 
    over 
    non-inclusive ones
    when translating from German into English. 
\end{enumerate}

The MuST-SHE$^{WMT23}$ and INES datasets, as well as their corresponding metrics and evaluations, are respectively discussed in Section \ref{sec:mustshe} and \ref{sec:INES}. 
%
%
In the evaluations presented in this paper, we obtained translations of our test suites by systems that are part of the standard General Translation Task of the Eighth Conference on Machine Translation (WMT-2023). In particular, we 
evaluated
11 systems for MuST-SHE$^{WMT23}$ en-de and 13 systems for INES de-en.

\section{MuST-SHE$^{WMT23}$: en-de Evaluation}
\label{sec:mustshe}

\begin{table*}[h!]
\centering
\footnotesize
\begin{tabular}{llp{11.2cm}c}
  \toprule
 \textbf{Form}
 &  & \textbf{Category 1}: \textit{Ambiguous first-person references} & \textbf{Speaker}\\ [1.mm]
 \cmidrule{2-3}
Fem. & \textsc{src} & The other hat that I've worn in my work is as \textbf{an activist}... & She \\
& \textsc{Ref}$_{De}$  &  Der andere Hut, den ich bei meiner Arbeit getragen habe, ist \textbf{der}\textbf{\scriptsize{<\texttt{den}>}} \textbf{Aktivistin}\textbf{\scriptsize{<\texttt{Aktivist}>}}... &\\ 
 \cmidrule{2-3}
Masc. & \textsc{src} & I mean, I'm a \textbf{journalist}. & He \\
& \textsc{Ref}$_{De}$ & Ich meine, ich bin \textbf{Journalist} \textbf{\scriptsize{<\texttt{Journalistin}>}}. &\\

 \midrule
  & &  \textbf{Category 2}: \textit{Unambiguous references disambiguated by gender info} &\\ [1.mm]
   \cmidrule{2-3}
   Fem. & \textsc{src} &     A college classmate wrote me a couple weeks ago and \underline{she} said ... & He\\ 
   & \textsc{Ref}$_{De}$ &  \textbf{Eine}\textbf{\scriptsize{<\texttt{Ein}>}} \textbf{Kommilitonin}\textbf{\scriptsize{<\texttt{Kommiliton}>}} hat mir vor ein paar Wochen geschrieben und gesagt...\\
\cmidrule{2-3}
Masc. & \textsc{src} & I decided to pay a visit to \textbf{the manager} [...] and \underline{he} pointed ...  &  She \\ 
& \textsc{Ref}$_{De}$ & Also entschied ich mich \textbf{den}\textbf{\scriptsize{<\texttt{die}>}}\textbf{Filialleiter}\textbf{\scriptsize{<\texttt{Filialleiterin}>}} zu besuchen [...] & \\
\midrule
\end{tabular}
\caption{MuST-SHE annotated segments organized per category. 
For each gender-neutral word referring to a human entity in the English source sentence (SRC), the reference translation (REF) shows the corresponding gender-marked (Fem./Masc.) forms, annotated with their wrong <gender-swapped> forms. 
The last column of the table provides information about the speaker's gender.}
\label{tab:Examples}
\end{table*}

MuST-SHE$^{WMT23}$ is a test suite designed to evaluate the ability of MT systems to correctly translate gender.
It is composed of 200 segments that require the translation of at least one English gender-neutral word into the corresponding masculine or feminine target word(s) in German.
The test suite is created as an extension of MuST-SHE,  a multilingual, natural benchmark built on TED talks data \citep{bentivogli-2020-gender}, which allows for a fine-grained analysis of gender bias in MT and ST. 
%
The original MuST-SHE corpus 
comprises $\sim$3,000 (\textit{audio, transcript, translation}) triplets annotated with qualitatively differentiated gender-related phenomena for thee language pairs: English$\rightarrow$ French/Italian/Spanish. 
Here, we introduce a newly created \textbf{English$\rightarrow$ German} textual portion (\textit{transcript, translation}) of the MuST-SHE corpus. 



\subsection{MuST-SHE$^{WMT23}$ Dataset}

\paragraph{Phenomena of Interest.}
Following the MuST-SHE original design, MuST-SHE$^{WMT23}$ is intended to evaluate
the translation of a source English neutral word into its corresponding target gender-marked one(s) in the context of human referents, e.g. en: \textit{\textbf{the} good \textbf{friend}}, de: \textit{\underline{\textbf{der}}/\underline{\textbf{die}} gute \textbf{Freund/\underline{in}}}.

To allow revealing a potential gap across the generation of  feminine/masculine gender forms, the corpus includes a balanced number of feminine (F) and masculine (M) translation phenomena.
%
Also, the corpus features two 
categories of phenomena, which differ in the 
presence/lack
of a gender cue to disambiguate the translation. Namely, \textit{i)} \textbf{CAT1}: consisting of first-person singular references (i.e. to the speaker), which are to be translated according to the speaker's linguistic expression of gender, e.g., \textit{I am a good friend}. 
Then, \textit{ii)} \textbf{CAT2} consisting of
references to any participant,  
which are be 
translated according to explicit gender information available in the sentence, like lexically gendered words (\textit{sister}, \textit{Mr}), or pronouns
(\textit{\underline{He/she} is a good friend}). These categories allow differentiating systems' behaviour across ambiguos vs. unambiguos cases. 

\paragraph{Dataset creation.}

In order to create MuST-SHE$^{WMT23}$ 
we collected a pool of English-German candidate segments by exploiting the same TED-based data sources used to create the other MuST-SHE datasets, namely: the Dev and Common Test sets of the MuST-C corpus, and other parallel sentences extracted from additional TED talks.
%
%
Then, to target those segments that represented our phenomena of interest, we followed the same automatic procedure used for the original MuST-SHE benchmark, which was aimed to quantitatively and qualitatively maximize the extraction of an assorted variety of gender-marked phenomena. 
Regular expressions were employed to transform German gender-agreement rules into search patterns to be applied to our pool of candidate sentences. 
Also, to specifically
match a differentiated range of gender-marked lexical items, we also compiled two series of 50 human-referring adjectives in English and German.

Once the automatic step was concluded, the pool of retrieved sentence pairs underwent a manual inspection to: \textit{i)} remove any noise and keep only pairs containing at least one gender phenomenon; \textit{ii)} ensure that the final (\textit{transcript, translation}) pairs were not affected by misalignments resulting from the automatic procedure used to create MuST-C and the new TED Talks data.  
Also, we examined the remaining pairs to verify that those to be included in MuST-SHE featured a 
 a balanced distribution of categories, F/M forms, and speakers. 
 Accordingly, since the MuST-C corpus presents a well-known gender imbalance\footnote{As reported in MuST-Speakers, $\sim$70\% of the speakers in MuST-C are referred to by \textit{He} pronouns.}, we 
excluded all of the extracted masculine segments that exceeded the feminine counterpart. 
Across categories, instead, we were not able to ensure a balanced distribution, as fewer instances from CAT1 could be identified.\footnote{This is most likely due to the gendered features of the German language, which -- unlike es, fr, and it -- does not carry gender markings on verbs (e.g., \textit{I} \textit{went} $\rightarrow$ de: \textit{Ich bin gegangen} vs it: \textit{Sono }\textit{andat\textbf{a}/\textbf{o}}) nor adjective in the nominative case (e.g., \textit{I am good} $\rightarrow$ de: \textit{Ich bin gut} vs. es: \textit{Soy buen\textbf{a/o}}.}

The resulting dataset -- whose statistics are given in Table \ref{tab:mustshe_statistics} -- was then manually enriched with different types of information. For each segment, the annotation includes: category (1 and 2), gender form (F and M), and
speaker’s gender information.\footnote{Such an information is migrated from the MuST-Speakers resource \citep{gaido-etal-2020-breeding}, where  gender information for each speaker in MuST-C has been labeled based on the personal pronouns the speakers used to describe themselves in their publicly available personal TED section.}
%
%
Also, for each target gender-marked word in MuST-SHE$^{WMT23}$,
we created a corresponding gender-swapped counterpart in the opposite gender form. 
As shown in Table \ref{tab:Examples}, 
these word forms were paired and annotated in the reference translations.
As we will describe in more detail in the upcoming Section \ref{subsec:mustshe_eval}, such annotated target gender-marked words are key features of MuST-SHE, which enable gender-sensitive, fine-grained analyses focusing solely on the correct generation of target gender-marked words. 

The manual selection of appropriate sentences and their annotation was carried out by two annotators, 
both students proficient in the German language and with a background in Applied Linguistics.\footnote{Their work was carried out as part of an internship at FBK.}
Each annotator worked on half of the corpus independently and then revised the work done by the other. Finally, all the differences found were reconciled
to get to the final corpus.



\begin{table}[htp]
\centering
\small
\begin{tabular}{@{}r|cc|@{}}
                         & \textbf{CAT1}       & \textbf{CAT2}       \\
\toprule
\textbf{Fem.}                      & 23         & 77         \\
\textbf{Masc.}                        & 23         & 77         \\
\midrule
\multicolumn{1}{r|}{\textbf{Tot.}} & \multicolumn{2}{c|}{200} \\
\bottomrule
\end{tabular}
\caption{MuST-SHE$^{WMT23}$ sentence-level statistics.}
\label{tab:mustshe_statistics}
\end{table}

\subsection{MuST-SHE$^{WMT23}$ Evaluation}
\label{subsec:mustshe_eval}

Following the original MuST-SHE evaluation protocol described in \citet{gaido-etal-2020-breeding}, MuST-SHE$^{WMT23}$ evaluation allows to focus on the gender realization of the target gender-marked forms, which are annotated in the reference translations together with their \textit{wrong}, gender-swapped form (see Table \ref{tab:Examples}). The evaluation is carried out in two steps, and by matching the annotated (\textit{correct}/\textit{wrong}) gender-marked words against the MT output.  
%
%
Accordingly,  we first 
 calculate the 
 \textbf{Term Coverage}
as the proportion of gender-marked words annotated in MuST-SHE (either in the correct or wrong form) that are actually generated by the system, on which the accuracy of gender realization is therefore \textit{measurable}. 
Then, we define 
\textbf{Gender Accuracy}
as the proportion
of correct gender realizations among the words on which it is \textit{measurable}.
This 
evaluation method\footnote{The evaluation script is publicly available at: \url{https://github.com/hlt-mt/FBK-fairseq/blob/master/examples/speech_to_text/scripts/gender/mustshe_gender_accuracy.py}.}
%
has several advantages.
On one side,
\textit{term coverage} unveils the precise amount of words on which systems' gender realization is measurable.
On the other, \textit{gender accuracy}  directly informs about systems' performance on gender translation and related gender bias: 
scores below 50\% indicate that the system produces the wrong gender more often than the correct one, thus signalling a 
particularly strong biased behaviour.

\begin{table*}[t]
\small
\centering
\renewcommand{\arraystretch}{1.1}
\begin{tabular}{@{}l||c|c|cccc@{}}
\textbf{} &
  \multicolumn{1}{c|}{\textbf{All-Cov}} &
  \multicolumn{1}{c|}{\textbf{All-Acc}} &
  \multicolumn{1}{c}{\textbf{1F-Acc}} &
  \multicolumn{1}{c}{\textbf{1M-Acc}} &
  \multicolumn{1}{c}{\textbf{2F-Acc}} &
  \multicolumn{1}{c}{\textbf{2M-Acc}} \\
  \hline
  \hline
\textit{ONLINE-M}        & 75.07          & 80.07 & 50.00 & 84.00           & 86.08          & 80.00           \\
\textit{ONLINE-Y}        & 73.35          & 79.65          & 30.43          & 96.15           & 86.96          & 78.51          \\
\textit{NLLB\_MBR\_BLEU} & 71.92          & 79.43          & 36.00          & 92.31           & 87.27          & 78.51          \\
\textit{ONLINE-W}        & 67.91          & 79.32          & 23.81          & 90.91           & 86.11          & 80.87          \\
\textit{ONLINE-G}        & 77.07 & 78.87          & 16.00          & 95.15           & 87.39 & 79.69          \\
\textit{ONLINE-B}        & 72.20           & 78.64          & 14.28          & 100.00 & 83.92          & 81.25 \\
\textit{ONLINE-A}        & 74.78          & 78.00             & 25.00             & 92.30            & 84.34          & 79.36          \\
\textit{GPT4-5shot}      & 69.63          & 77.94          & 10.53          & 95.83           & 83.33          & 80.17          \\
\textit{ZenhHuiMT}       & 73.07          & 77.35          & 19.23          & 95.65           & 84.82          & 79.37    \\
\textit{Lan-BridgeMT}    & 71.92          & 75.79          & 16.67          & 92.31           & 83.19          & 77.05          \\
\textit{AIRC}            & 67.34          & 73.98          & 10.53          & 87.50            & 81.25          & 74.56          \\
\hline
\end{tabular}
\caption{MuST-SHE$^{WMT23}$ results for en-de. Systems are ranked based on overall Gender Accuracy (All-Acc).}
\label{tab:mustshe_resulsts}
\end{table*}

\subsection{MuST-SHE$^{WMT23}$ Results}
\label{must-she_results}

In Table \ref{tab:mustshe_resulsts} we present the MuST-SHE$^{WMT23}$ results 
for the 11 en-de systems that
were submitted to the WMT-2023 standard General Translation Task. 
Starting from coverage results, 
the scores range between 67.34\% (\texttt{AIRC}) and 77.07\%  (\texttt{ONLINE-G}), with only 3 systems under 70\%.
Hence, overall all models produce a good amount of gender-marked words that 
can  be evaluated with regards to the accuracy of their gender realization.
%
%
Moving onto the overall accuracy scores (All-Acc), we can see that -- while there is still room for improvement -- all of the evaluated MT systems are reasonably good at translating gender, 
with \texttt{ONLINE-M} emerging as the best model, able to correctly translate gender in 80\% of the generated instances. If we go more fine-grained into results disaggregated across gender forms (F and M) and categories (1 and 2), however, we can unveil subtle differences. Indeed, for unambiguous gender translation from CAT2, systems perform basically on par across gender forms, with actually slightly higher results for feminine translation. Instead, results on CAT1 unveil a huge gender gap, with systems achieving almost perfect results for masculine translation, whereas feminine accuracy can be as low as 10.53\%. In fact, the best ranked systems \texttt{ONLINE-M} generates the correct feminine form in 50\% of the cases, namely at a random rate. 

Overall, results on MuST-SHE$^{WMT23}$ show that the evaluated MT systems are reasonably good at translating gender under realistic conditions, achieving comparable results across feminine and masculine gender translation. However, for ambiguous cases where the input sentence does not inform about the gender form to be used in translation, we confirm a strong skew where all systems favour masculine generation almost by default.
This finding calls for further research endeavours and evaluation initiatives to counter gender bias in MT and measure future advances.

\section{INES: de-en Evaluation}
\label{sec:INES}

\begin{table*}[t]
\small
\begin{tabular}{@{}lll@{}}
\toprule
 & German src & English pair \\ \midrule
a. & \textbf{Der Polizist} half der alten Dame, die Straße sicher zu überqueren. & \textbf{police officer}, \textbf{policeman} \\
b. & \textbf{Die Menschheit} hat das Potenzial, die Welt zu einem besseren Ort zu machen. & \textbf{humankind}, \textbf{mankind} \\
c. & Die \textbf{fachmännische} Arbeit des Teams führte zum erfolgreichen Abschluss des Projekts. & \textbf{skillful}, \textbf{workmanlike} \\
d. & Die \textbf{geschickte} Arbeit des Teams führte zum erfolgreichen Abschluss des Projekts. & \textbf{skillful}, \textbf{workmanlike} \\ \bottomrule
\end{tabular}
\caption{INES source German example sentences with their corresponding annotated English \texttt{IN} and \texttt{N-IN} terms.}
\label{tab:INES_sentences}
\end{table*}



The INclusive Evaluation Suite (INES) is a test set designed to assess MT systems ability to produce gender-inclusive translations for the German$\rightarrow$English language pair.
By design, each German source sentence in INES includes an expression that
can be rendered by means of either an \textit{inclusive} (\texttt{IN}) or 
\textit{non-inclusive}
(\texttt{N-IN)} expression in the English target language. 
%

%
Overall, INES comprises 162 manually curated German sentences, which are annotated with their corresponding (\texttt{IN/N-IN}) English expressions. As such, it allows to evaluate to what extent MT systems favor the generation of 
non-inclusive
solutions over alternative, valid inclusive translation in their output. 

\subsection{INES Dataset}
Here, we first describe the phenomena of interest included in INES. Then, we proceed by describing its creation methodology. 

\paragraph{Phenomena of interest.}
Despite its comparatively restricted gender grammar, English has traditionally relied on the use of marked forms that treat the masculine gender as the conceptually generic, default human prototype, i.e. as \textit{masculine generics} \citep{silveira1980generic, bailey2022based}.
Exemplary cases of such a phenomenon are man-derivates  (e.g., \textit{man-made}, \textit{freshman}) and the use of masculine personal pronouns for generic referents (e.g., ``each student must submit \textit{his} form''). Besides, expressions such as ``\textit{man} and \textit{wife}'' have been identified as depicting skewed representation of genders and gender roles \citep{stahlberg2007representation}. 
Toward the adoption of fairer language for all genders,
alternative and inclusive solutions are increasingly promoted by institutions \citep{hoglund2023gendering} and recommended in writing \citep{apa2020publication}. These include the use of unmarked forms (e.g. \textit{human-made}, \textit{first-year student}) and neutral pronouns (e.g. ``each student must submit \textit{their} form'') for generic and under-specified referents, as well as more symmetrical formulations that cast men and women in the same role (e.g.  ``\textit{husband} and \textit{wife}''). 

On this basis,  INES 
represents translation phenomena where, given a source German sentence, systems are confronted with the generation of a corresponding inclusive or 
non-inclusive
solution. 
As shown by the examples in Table \ref{tab:INES_sentences}, the
German sentences can entail the use of either  \textit{i)}  a generic masculine form,
e.g. \textit{Der Polizist}, or \textit{ii)} 
a term that does not convey gender, e.g. \textit{Die Menschheit}. 
Regardless of the source German term, the expected ideal behaviour of the MT system always entails the generation of inclusive target words.

\paragraph{Dataset Creation.}
Since the focus of the INES test suite is to evaluate the ability of MT systems to generate inclusive English translations, 
we started by compiling a list of 
well-established
pairs of English \texttt{IN/N-IN} terms and expressions.
This list was created based on existing collections of paired terms \citep{vanmassenhove-etal-2021-neutral,amrhein-etal-2023-exploitings} and integrated with few additional terms retrieved from other inclusive language guidelines from international institutions\footnote{\url{https://www.europarl.europa.eu/cmsdata/151780/GNL_Guidelines_EN.pdf}} and universities.\footnote{\url{https://writingcenter.unc.edu/tips-and-tools/gender-inclusive-language/.}}\footnote{\url{https://www.gsws.pitt.edu/resources/faculty-resources/gender-inclusive-non-sexist-language-guidelines-and-resources}.} As a result, we obtained 48 \texttt{IN/N-IN} English pairs, which are 
shown in Table \ref{tab:INES_pairs}.

\begin{table}[!h]
\centering
\scriptsize
\begin{tabular}{cc}
\toprule
\multicolumn{2}{c}{\textbf{IN vs N-IN for job titles}}                \\
\midrule
\multicolumn{1}{c}{anchor}            & anchorman            \\
\multicolumn{1}{c}{anchors}           & anchormen            \\
\multicolumn{1}{c}{bartender}         & barman               \\
\multicolumn{1}{c}{bartenders}        & barmen               \\
\multicolumn{1}{c}{business person}   & businessman          \\
\multicolumn{1}{c}{business persons}  & businessmen          \\
\multicolumn{1}{c}{chairpeople}       & chairmen             \\
\multicolumn{1}{c}{chairperson}       & chairman             \\
\multicolumn{1}{c}{firefighter}       & fireman              \\
\multicolumn{1}{c}{firefighters}      & firemen              \\
\multicolumn{1}{c}{flight attendant}  & steward              \\
\multicolumn{1}{c}{flight attendants} & stewards             \\
\multicolumn{1}{c}{mail carrier}  & postman              \\
\multicolumn{1}{c}{mail carriers} & postmen             \\
\multicolumn{1}{c}{member of congress} & congressman \\
\multicolumn{1}{c}{members of congress} & congressmen \\
 \multicolumn{1}{c}{police officer} & policeman \\
    \multicolumn{1}{c}{police officers} & policemen \\
    \multicolumn{1}{c}{principal} & headmaster \\
    \multicolumn{1}{c}{principals} & headmasters \\
    \multicolumn{1}{c}{salesperson} & salesman \\
    \multicolumn{1}{c}{salespersons} & salesmen \\
\multicolumn{1}{c}{spokesperson}      & spokesman            \\
    \multicolumn{1}{c}{spokespeople} & spokesmen \\
\multicolumn{1}{c}{supervisor}        & foreman              \\
\multicolumn{1}{c}{supervisors}       & foremen              \\
\multicolumn{1}{c}{weather reporter}  & weatherman           \\
\multicolumn{1}{c}{weather reporters} & weathermen           \\
\toprule
\multicolumn{2}{c}{\textbf{IN vs N-IN for generic man}}      \\
\midrule
\multicolumn{1}{c}{average person}    & average man          \\
\multicolumn{1}{c}{average people}    & average men          \\
best people for the job               & best men for the job \\
best person for the job               & best man for the job \\
human-made                            & man-made             \\
humankind                             & mankind              \\
husband and wife                      & man and wife         \\
intermediaries                        & middlemen            \\
intermediary                          & middleman            \\
skillful                              & workmanlike          \\
laypeople                             & laymen               \\
layperson                             & layman               \\
workforce                             & manpower             \\
first-year student                    & freshman             \\
first-year students                   & freshmen             \\
\toprule
\multicolumn{2}{c}{\textbf{IN vs N-IN pronouns}}             \\
\midrule
their                                 & his                  \\
theirs                                & his                  \\
them                                  & him                  \\
themself                              & himself              \\
they                                  & he            \\
\bottomrule
\end{tabular}
\caption{INES pairs of English Inclusive (\texttt{IN}) vs 
non-inclusive
(\texttt{N-IN}) expressions.}
\label{tab:INES_pairs}
\end{table}

Starting from this list, we created the source German sentences that compose INES following a two-step semi-automatic procedure. 

In the first step, for each English \texttt{IN/N-IN} term
of
the pairs, GPT\footnote{\texttt{gpt-3.5-turbo}.} was prompted to
generate 3 German sentences containing such
term translated into German, for a total of 6 sentences for each English pair.
  
In the second step, the initial pool of 288 synthetic sentences was manually revised by a linguist proficient in German.\footnote{One of the authors of the paper.} The revision was aimed to \textit{i)} correct 
generation errors and 
\textit{ii)} select a balanced amount of German sentences for each phenomenon of interest. To this purpose: 
\begin{itemize}
    \item when all the 6 German sentences generated for the two (\texttt{IN/N-IN}) terms of the English pair contained only gender-marked terms (e.g. \textit{police officer} --> \textit{Der Polizist} / \textit{policeman} --> \textit{Der Polizist}) or only gender-neutral terms (e.g. \textit{humankind} --> \textit{Die Menschheit} / \textit{mankind} --> \textit{Die Menschheit}), only 3 sentences out of 6 were kept (see examples a. and b. in Table \ref{tab:INES_sentences});
    \item on the contrary, when the 6 German 
    sentences generated for 
    the two (\texttt{IN/N-IN}) English terms included both gender-marked and gender-neutral forms (e.g. \textit{firefighters} --> \textit{Feuerwehrleute} / \textit{firemen} --> \textit{Feuerwehrmänner}), they were all kept, so as to have a richer representation of the phenomenon of interest in the source (see c. and d. in Table \ref{tab:INES_sentences}). 
\end{itemize}
 
Unfortunately, we found only very few instances of double German realizations, and thus at the end of the manual revision, we remained with 162 German sentences: 21 with an inclusive source term, and 141 with a non-inclusive masculine generic in the source. All the 162 manually-curated German source sentences are included in INES, and provided with their corresponding English \texttt{IN}/\texttt{N-IN} term pair so as to allow for focused evaluations. 

\subsection{INES Evaluation}
\label{subsec:ines_eval}

To evaluate systems against INES, we can leverage the annotated pairs of English \texttt{IN/N-IN} expressions and match them against the MT generated output. 
Accordingly, we can perform our evaluation by adopting  the same evaluation protocol and metrics defined for MuST-SHE in \ref{subsec:mustshe_eval}. Namely, by  \textit{i)} first computing \textbf{Term Coverage} as the proportion of \texttt{IN/N-IN} generated by a system, and then \textit{ii)} calculating \textbf{Inclusivity Accuracy} as the proportion of  \texttt{IN} generated expressions, among all of the generated ones. As a result, all the \textit{out of coverage words} (OOC) are necessarily left unevaluated. 

While prior manual assessments of the 
terms left unevaluated by such an automatic method
have been able to confirm the robustness and validity of the accuracy results in the context of binary gender translation \citep{savoldi-etal-2022-morphosyntactic}, here we hypothesise a potential limit for evaluating inclusivity in English outputs. 
Our hypothesis lies on the fact that English, a notional gender language \citep{McConnell-Ginet2013}, has a restricted repertoire of gender-marked -- potentially \texttt{N-IN} -- words, whereas most English nouns simply do not convey any gender distinctions (e.g., \textit{doctor}, \textit{secretary}, \textit{president}). In other words, there might be many potential inclusive alternatives and synonyms (e.g. \textit{presenter} and \textit{host} for \textit{<anchor>}) for a single  
\texttt{N-IN} term (e.g. \textit{<anchorman>}). Thus, whereas OOC words in the context of binary gender present the same distribution assessed automatically in terms of accuracy,  this metric might be stringent for inclusivity in English, and overly penalize the generation of alternative terms
that differ from those annotated in INES. 

%
%
%

In light of the above, we also propose the \textbf{Inclusivity Index} metric, defined as:

\begin{equation}
    \text{Inclusivity Index} = 1 - \frac{n_{\text{N-IN}}}{N}
\end{equation}

where $n_{\text{N-IN}}$ is the number of non-inclusive terms annotated in INES that are generated by a system, and $N$ is the size of INES (i.e. total number of sentences to be evaluated).

In what follows, we thus carry out both \textbf{Inclusivity Accuracy} and \textbf{Inclusivity Index} evaluations,\footnote{Evaluation script available at: \url{https://github.com/hlt-mt/FBK-fairseq/blob/master/examples/speech_to_text/scripts/gender/INES_eval.py}.} and assess which one better correlates with human judgments.

\begin{table*}[htbp]
  \centering
  \small
  \begin{subtable}{0.33\linewidth}
    \centering
    \begin{tabular}{@{}l||r|r@{}}
      \toprule
      & \textbf{Cov} & \textbf{Acc ($\uparrow$)} \\ \midrule
      \textit{GPT4-5shot}      & 64.81          & \textbf{65.71} \\
      \textit{ONLINE-W}        & \textbf{75.31} & 48.36          \\
      \textit{ONLINE-Y}        & 74.07          & 45.83          \\
      \textit{ZenhHuiMT}       & 73.46          & 44.54          \\
      \textit{ONLINE-A}        & 74.69          & 42.98          \\
      \textit{ONLINE-B}        & 70.99          & 41.74          \\ 
      \textit{AIRC}            & 53.70          & 37.93          \\
      \textit{Lan-BridgeMT}    & 68.52          & 36.94          \\
      \textit{ONLINE-M}        & 70.37          & 36.84          \\
      \textit{ONLINE-G}        & 74.07          & 35.00          \\
      \textit{GTCOM\_Peter}             & 74.69          & 33.06          \\
      \textit{NLLB\_Greedy}             & 74.07          & 31.67          \\
      \textit{NLLB\_MBR\_BLEU} & 73.46          & 29.41          \\
      \bottomrule
    \end{tabular}
        \caption{Coverage and Accuracy results}
    \label{subtab:cov-acc}
  \end{subtable}
  \begin{subtable}{0.33\linewidth}
    \centering  
    \begin{tabular}{@{}l||r@{}}
     \toprule
      & \textbf{In.Idx. ($\uparrow$)} \\ \midrule
      \textit{GPT4-5shot}            & 77.78                                \\
      \textit{AIRC}                  & 66.67                                \\
      \textit{ONLINE-W}              & 61.11                                \\
      \textit{ONLINE-Y}              & 59.88                                \\
      \textit{ZenhHuiMT}             & 59.26                                \\
      \textit{ONLINE-B}              & 58.64                                \\
      \textit{ONLINE-A}              & 57.41                               \\
      \textit{Lan-BridgeMT}          & 56.79                                \\
      \textit{ONLINE-M}              & 55.56                                \\
      \textit{ONLINE-G}              & 51.85                                \\
      \textit{GTCOM\_Peter}          & 50.00                                \\
      \textit{NLLB\_Greedy}          & 49.38                               \\
      \textit{NLLB\_MBR\_BLEU}       & 48.15                                \\
      \bottomrule
    \end{tabular}
     \caption{Inclusivity Index results}
    \label{subtab:inclusivity}
  \end{subtable}
  \begin{subtable}{0.33\linewidth}
    \centering
    \begin{tabular}{@{}l||r@{}}
    \toprule
      & \textbf{Human ($\uparrow$)} \\ \midrule
      \textit{GPT4-5shot}            & 76.73                             \\
      \textit{ONLINE-W}              & 60.25                             \\
      \textit{AIRC}                  & 59.03                             \\
      \textit{ONLINE-Y}              & 58.13                             \\
      \textit{ZenhHuiMT}             & 56.60                             \\
      \textit{ONLINE-B}              & 56.25                             \\
      \textit{ONLINE-A}              & 55.28                             \\
      \textit{ONLINE-M}              & 52.53                             \\
      \textit{Lan-BridgeMT}          & 52.26                             \\
      \textit{ONLINE-G}              & 48.45                             \\
      \textit{NLLB\_MBR\_BLEU}       & 46.25                             \\
      \textit{GTCOM\_Peter}          & 48.13                             \\
      \textit{NLLB\_Greedy}          & 44.03                             \\
      \bottomrule
    \end{tabular}
     \caption{Human judgment -- \textit{Official ranking}}
    \label{subtab:human-judgment}
  \end{subtable}
    \caption{INES evaluation results (percentage). Per each metric, systems are ranked based on their performance.}
    \label{tab:ines_results}
  \end{table*}


\begin{figure*}[t]
  \centering
\includegraphics[scale= 0.50]{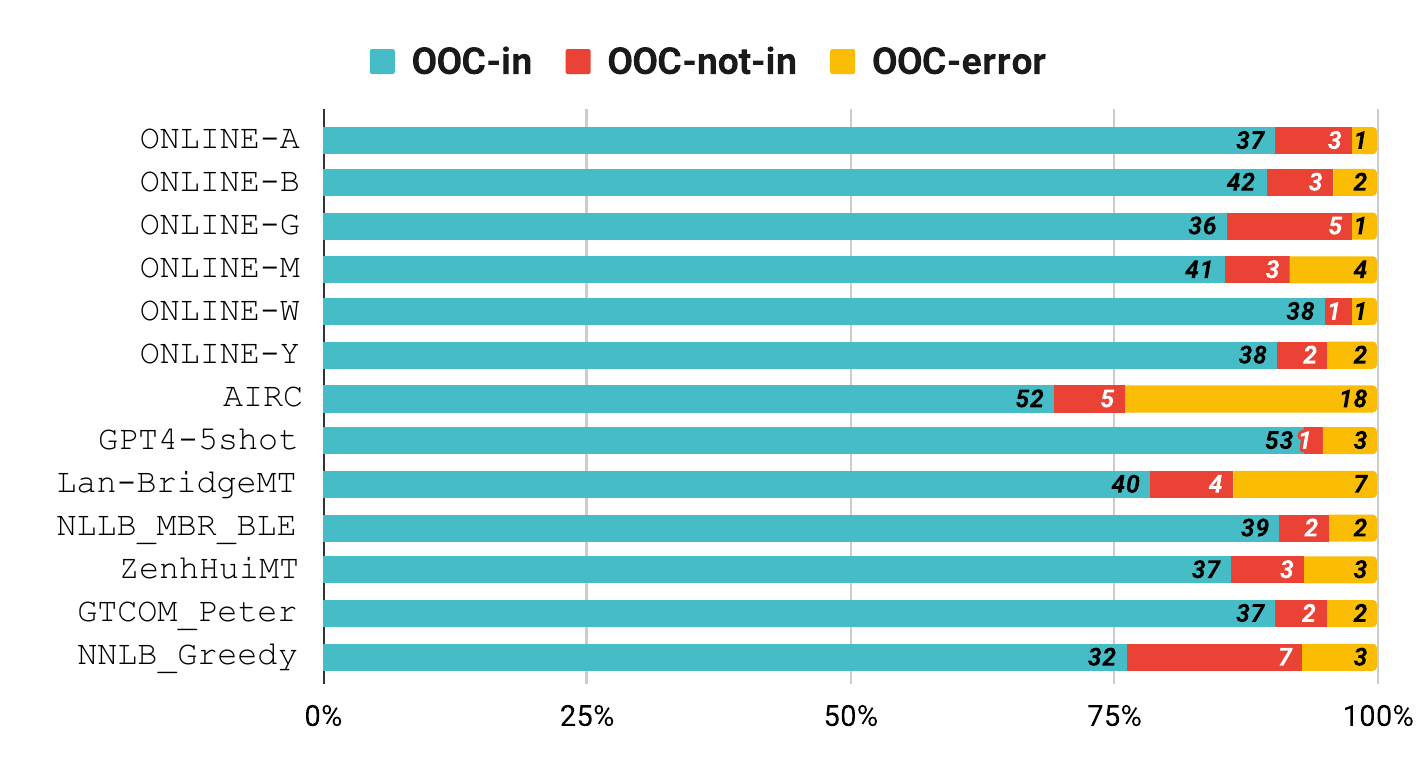}
  \caption{INES manual analysis results for out-of-coverage (OOC) terms.}
  \label{fig:ines-occ-percent}
\end{figure*}



\begin{table}[htbp]
  \centering
  \small
  \label{tab:correlation-coefficients}
  \begin{tabular}{@{}lcccc@{}}
    \toprule
    \textbf{Metric} & \textbf{Pearson} (r) & \textbf{Kendall ($\tau$)} & \textbf{Spearman ($\rho$)} \\
    \midrule
    Acc & 0.9601 & 0.8205 & 0.9285 \\
    In.Idx. & \textbf{0.9738} & \textbf{0.9231} & \textbf{0.9835} \\
    \bottomrule
  \end{tabular}
  \caption{Correlation Coefficients with Human Judgment}
  \label{tab:correlations}
\end{table}


\subsection{INES Results}
In this section (Table \ref{tab:ines_results}), we present the results obtained on INES by the 13 de-en systems that were submitted to the WMT-2023 standard General Translation Task. 
Such results are computed and discussed for Inclusivity Accuracy (Table \ref{subtab:cov-acc})
and
Inclusivity Index (Table \ref{subtab:inclusivity}). Then, based on a manual analysis, we compare such automatic results against the  systems ranking  obtained with human evaluations (Table \ref{subtab:human-judgment}). 

\paragraph{Automatic Evaluation Results.}
Table \ref{subtab:cov-acc} presents coverage and accuracy-based results. Based on such scores, the INES dataset emerges as quite a challenging test suite for current de-en systems. In fact, with the sole exception of the \textsc{GPT4-5shot}  -- which emerges as the best performing system (but see also Sec. \ref{sec:limitations}) -- all systems obtain scores that are below 50\%, thus suggesting that they generate undesirable \texttt{N-IN} forms in more than half of the (measurable) cases. The lowest accuracy is obtained by \textsc{NLLB\_MBR\_BLEU}, amounting to 29.41\% only. 

Moving onto the Inclusivity Index results in Table \ref{subtab:inclusivity}, from a bird's eye view we can immediately unveil some differences. On the one hand, \textsc{GPT4-5shot} and \textsc{NLLB\_MBR\_BLEU}  still emerge as, respectively, the best and worst performing systems. On the other hand, however, there are discrepancies in the overall ranking. For instance, \textsc{AIRC} results as the system that generates the second-best level of inclusive translation according to the Inclusivity Index metrics, whereas it was ranked 7th in terms of accuracy.

\paragraph{Manual Evaluation Results.}
To verify which of the two automatic metrics yields more reliable results, we proceed with a manual analysis of all 
MT output sentences that defied the automatic evaluation procedure. Namely, we performed a human evaluation of all OOC terms to determine whether the generated expression entailed \textit{i)} an inclusive 
expression
(OOC-in), which simply differed from the \texttt{IN} term annotated in INES but was completely acceptable; \textit{ii)} a 
non-inclusive expression
(OOC-not-in) different
from the \texttt{N-IN} term annotated in INES; and finally \textit{iii)} a translation error (OOC-error), which was not possible to judge in terms of inclusivity.\footnote{We underscore that such an analysis only concerns the terms representing the phenomena of our interest, whereas the overall judgement of the whole sentence is not accounted for.} The results of such an analysis across all systems are reported in Figure \ref{fig:ines-occ-percent}. Such results show 
that, of all the OOC terms, the vast majority 
is represented by inclusive terms (e.g., <\textit{business person}>/<\textit{busissnessman}>$\rightarrow$ \textit{entrepeneur}). Errors, instead, are quite rare, just like 
non-inclusive
OOC terms, which all correspond to 
the INES annotated \texttt{N-IN} term, but in a different number 
(e.g., \textit{<freshmen>} $\rightarrow$ \textit{freshman}). 

In light of the above, our initial hypothesis -- outlined in Sec. \ref{subsec:ines_eval} -- is thus reinforced: we do not find the same inclusivity distribution between evaluated cases in terms of accuracy (see Table \ref{subtab:cov-acc}) and the OCC instances left unevaluated. Having now collected a complete evaluation of all the sentences,
we leverage such information to  
obtain our 
official system ranking, which is shown in Table \ref{subtab:human-judgment}.
Results 
are computed as the proportion of inclusive (\texttt{IN} + OOC-in) terms generated by a system among all the terms that could be assessed 
(i.e. OOC-errors are not measurable, hence excluded). 

\paragraph{Correlation between Automatic and Human evaluation.} On this basis, and to finally verify our hypothesis, in Table \ref{tab:correlations} we report the correlation coefficients between the automatic metrics and human 
judgements.
Accordingly, while both the Inclusivity Accuracy and Index show a satisfactory correlation with human judgements, the latter consistently emerges as a more reliable indicator of inclusivity. As such, Inclusivity Index is confirmed as the most suited measure to quantify gender-inclusive translation into English. 

\smallskip

To conclude, our results in Tables \ref{tab:ines_results} consistently indicate that current MT systems still struggle with the generation of inclusive translations. Within this landscape, \textsc{GPT4-5shot} consistently results as the model achieving the highest level of inclusivity, whereas all other models generate a $\sim$40\% or more of non-inclusive translations in their output. 
This finding highlights that, while on the (binary) gender \textit{bias} side (Section \ref{must-she_results}) MT systems still struggle with specific and particularly challenging ambiguous cases, the limitations of most of them on the gender \textit{inclusion} side are evident and the problem emerges as an urgent topic for future research.


\section{Related work}
\label{sec:related_work}

The last few years have witnessed and increasing attention toward (binary) gender bias in NLP \citep{sun-etal-2019, stanczak2021survey, savoldi-etal-2022-dynamics}.
Concurrently, emerging research has 
highlighted the importance of reshaping 
gender in NLP technologies in a more inclusive 
manner  \cite{dev-etal-2021-harms},
also through the representation of non-binary identities in 
language  \citep{lauscher-etal-2022-welcome,ovalle2023m}.
Foundational works in this area have included several applications, such as coreference resolution systems \citep{cao-daume-iii-2020-toward, brandl-etal-2022-conservative} and 
fair rewriters \citep{vanmassenhove-etal-2021-neutral,amrhein-etal-2023-exploitings}.  

In MT, the research agenda has mainly focused on the
improvement 
of masculine/feminine gender translation 
into grammatical gender languages \citep{savoldi2021gender}.
Along this line, different strategies have been devised to improve gender translation and mitigate masculine bias \citep{costa-jussa-de-jorge-2020-fine,gaido-etal-2021-split, choubey2021improving, saunders-etal-2022-first}.
To test these methods and inspect systems' behaviour, several  template-based datasets have been made available -- 
such as WinoMT \citep{stanovsky-etal-2019-evaluating} or SimpleGEN \citep{renduchintala-williams-2022-investigating} -- which are especially intended to target occupational stereotyping. 
Instead, natural datasets such as the Arabic Parallel Gender Corpus \citep{alhafni-etal-2022-user} and GATE \citep{rarrick2023gate} allow for evaluation of gender bias under more naturalistic conditions. Among such type corpora, MuST-SHE \citep{bentivogli-2020-gender} represents the only multilingual, natural test set designed to evaluate gender bias for both MT and ST. Already available for English$\rightarrow$French/Italian/Spanish, here we have contributed to its expansion for the English$\rightarrow$German language pair. 

As far as the topic of inclusivity and neutral language translation is concerned, research in MT is quite in its infancy. A notable exception is the work by \citet{saunders-etal-2020-neural},
who created
parallel test and fine-tuning data to
develop MT systems able to generate
non-binary translations 
for English$\rightarrow$German/Spanish. However,
their 
target sentences are artificial -- created by replacing gendered morphemes and articles with synthetic 
placeholders -- thus serving only as a proof-of-concept. \citet{piergentili-etal-2023-gender}, instead,  are the first to advocate for the use of target gender-neutral rephrasings and synonyms as a viable paradigm toward more inclusive MT when gender is unknown or simply irrelevant. \citet{cho-etal-2019measuring} and \citet{ghosh2023chatgpt}
investigate the preservation of gender-neutral pronouns
for Korean/Bengali$\rightarrow$English. Their results, however, show that current MT systems still face serious difficulties on relying on the inclusive, neutral pronoun \textit{they} in translation. Along this line of work, INES -- to the best of our knowledge -- represents the first test suite designed to asses the use of neutral, inclusive forms beside pronouns for translating into English.

\section{Conclusion}

This paper summarizes the results of our WMT-2023
Test Suites evaluations, which focus on gender bias and inclusivity in translation. 
To this aim, we have introduced the en-de expansion of the multilingual MuST-SHE test set \citep{bentivogli-2020-gender} and the newly created INES dataset for de-en.  
The former is designed to assess gender bias and translation across a qualitatively differentiated selection of feminine/masculine gender phenomena. INES, instead, measures systems' ability to generate inclusive English translations that do not rely on the use of masculine generics.
Results on MuST-SHE$^{WMT23}$ show that the evaluated MT systems are reasonably good at translating gender under realistic conditions, achieving comparable results across feminine and masculine gender translation. However, for ambiguous cases where the input sentence does not inform about the gender form to be used in translation, we confirm a strong skew where all systems tend to generate masculine forms almost by default. Results on INES, instead, indicate that providing inclusive translations still represents a quite challenging task for current MT systems, in spite of the increasingly widespread use and preference for inclusive language forms in English.

As a final remark, we acknowledge that the phenomena subject to our analysis (gender bias and gender inclusion) are not yet part  of the repertoire of phenomena for which MT systems are currently designed. These systems are indeed primarily built with the goal of maximising translation quality in general rather than aspects of the problem, specifically fairness, for which sensitivity is still limited.
All in all, however, this experience has allowed us to shed light on these issues, raise the awareness of the MT community and, hopefully, favour future developments.

\section*{Limitations}
\label{sec:limitations}
Naturally, this work comes with some limitations.
First, both test suites are limited 
in
size and number of language pairs considered. 
Despite their restricted size, however, both test suites provide a first glimpse into understanding and monitoring systems' behaviour with respect to gender and inclusivity. 
Additionally, rather than a limitation per se, both INES and MuST-SHE$^{WMT23}$ are designed based on the specific linguistic features of the source and target language taken into account. As such, results in our evaluations intentionally do not aspire to scale and generalize to any language direction. Indeed, such 
linguistic specificity is also openly accounted for in the introduction of the new Inclusivity Index metric, which considers the morphology of English for a better-suited evaluation of 
gender inclusivity in MT. 
We also note that such a metric results as the best one for evaluating inclusivity under the given experimental conditions of this paper, where all the 
scrutinized systems (those
submitted to the WMT General Translation task) are expected to 
feature
generally good overall translation quality and to make few translation errors. As such, future work might be needed to further validate the stability of the Inclusivity Index metric under less optimal conditions and for different target languages, 
possibly
proposing tailored metrics for each case.
Finally, to generate 
the initial pool of sentences in INES
we relied on the GPT (gpt-3.5-turbo) closed-source model. 
This has holds two types of implications. On the one hand, the use of proprietary models such as GPT 
has reproducibility consequences, since this model is regularly updated, thus potentially yielding future results that differ from those reported in this paper. 
On the other hand, relying on -- even though only partially and post-edited -- artificially generated data for testing models, might lead to contamination issues. Indeed, in Sec. \ref{subsec:ines_eval} (Table \ref{tab:ines_results}) the \textsc{GPT4-5shot} model resulted as the most promising one, achieving the best results for inclusive translation. However, it remains to further verified whether our specific experimental settings and INES benchmark -- where we use GPT-derived test data --  have advantaged the performance of \textsc{GPT4-5shot}.

\section*{Ethics Statement}

By addressing bias and inclusivity in MT, this work presents an inherent ethical component. It builds from concerns toward the societal impact of widespread translation technologies that reflect and propagate male-grounded and exclusionary language.
Still, our work is not without risks either and thus warrants some ethical considerations. In particular, MuST-SHE$^{WMT23}$ only focuses on traditional binary feminine/masculine gender forms. Also, INES investigates neutral, inclusive language in the context of generic, unknown referents and based on inclusive solutions encouraged by institutional guidelines. As such, we do not account for other non-binary solutions (e.g., neopronouns and neomorphemes) that are 
emerging from
grassroots efforts. 
It should be stressed that the gendered and inclusive strategies incorporated in this MT work are not prescriptively intended. Rather, they are orthogonal to other attempts and non-binary expressions for inclusive language (technologies) \citep{lauscher2023em, rivas-ginel-2022-all-inclusive}.

\section*{Acknowledgements}

This work is part of the project ``Bias Mitigation and Gender Neutralization Techniques for Automatic Translation”, which is financially supported by an Amazon Research Award AWS AI grant. 
Also, we acknowledge the support of the PNRR project FAIR -  Future AI Research (PE00000013),  under the NRRP MUR program funded by the NextGenerationEU. 
Also, we would like to thank the 2022 FBK internship students
Sabrina Raus and Abess Benissmail from the University of Bolzano: the creation of MuST-SHE$^{WMT23}$ was made possible by their work. 

\bibliography{custom,anthology}

\begin{thebibliography}{36}
\expandafter\ifx\csname natexlab\endcsname\relax\def\natexlab#1{#1}\fi

\bibitem[{Alhafni et~al.(2022)Alhafni, Habash, and
  Bouamor}]{alhafni-etal-2022-user}
Bashar Alhafni, Nizar Habash, and Houda Bouamor. 2022.
\newblock \href {https://doi.org/10.18653/v1/2022.naacl-main.46} {User-centric
  gender rewriting}.
\newblock In \emph{Proceedings of the 2022 Conference of the North American
  Chapter of the Association for Computational Linguistics: Human Language
  Technologies}, pages 618--631, Seattle, United States. Association for
  Computational Linguistics.

\bibitem[{Amrhein et~al.(2023)Amrhein, Schottmann, Sennrich, and
  L{\"a}ubli}]{amrhein-etal-2023-exploitings}
Chantal Amrhein, Florian Schottmann, Rico Sennrich, and Samuel L{\"a}ubli.
  2023.
\newblock \href {https://doi.org/10.18653/v1/2023.acl-long.246} {Exploiting
  biased models to de-bias text: A gender-fair rewriting model}.
\newblock In \emph{Proceedings of the 61st Annual Meeting of the Association
  for Computational Linguistics (Volume 1: Long Papers)}, pages 4486--4506,
  Toronto, Canada. Association for Computational Linguistics.

\bibitem[{APA(2020)}]{apa2020publication}
APA. 2020.
\newblock \emph{Publication Manual of the American Psychological Association},
  7th edition.
\newblock American Psychological Association.

\bibitem[{Bailey et~al.(2022)Bailey, Williams, and Cimpian}]{bailey2022based}
April~H Bailey, Adina Williams, and Andrei Cimpian. 2022.
\newblock Based on billions of words on the internet, people= men.
\newblock \emph{Science Advances}, 8(13):eabm2463.

\bibitem[{Bentivogli et~al.(2020)Bentivogli, Savoldi, Negri, Di~Gangi, Cattoni,
  and Turchi}]{bentivogli-2020-gender}
Luisa Bentivogli, Beatrice Savoldi, Matteo Negri, Mattia~A. Di~Gangi, Roldano
  Cattoni, and Marco Turchi. 2020.
\newblock \href {https://www.aclweb.org/anthology/2020.acl-main.619} {{Gender
  in Danger? {E}valuating {S}peech {T}ranslation {T}echnology on the
  {M}u{ST}-{SHE} {C}orpus}}.
\newblock In \emph{Proceedings of the 58th Annual Meeting of the Association
  for Computational Linguistics}, pages 6923--6933, Online. Association for
  Computational Linguistics.

\bibitem[{Blodgett et~al.(2020)Blodgett, Barocas, Daum{\'e}~III, and
  Wallach}]{blodgett-etal-2020-language}
Su~Lin Blodgett, Solon Barocas, Hal Daum{\'e}~III, and Hanna Wallach. 2020.
\newblock \href {https://doi.org/10.18653/v1/2020.acl-main.485} {Language
  (technology) is power: A critical survey of {``}bias{''} in {NLP}}.
\newblock In \emph{Proceedings of the 58th Annual Meeting of the Association
  for Computational Linguistics}, pages 5454--5476, Online. Association for
  Computational Linguistics.

\bibitem[{Brandl et~al.(2022)Brandl, Cui, and
  S{\o}gaard}]{brandl-etal-2022-conservative}
Stephanie Brandl, Ruixiang Cui, and Anders S{\o}gaard. 2022.
\newblock \href {https://doi.org/10.18653/v1/2022.naacl-main.265} {How
  conservative are language models? adapting to the introduction of
  gender-neutral pronouns}.
\newblock In \emph{Proceedings of the 2022 Conference of the North American
  Chapter of the Association for Computational Linguistics: Human Language
  Technologies}, pages 3624--3630, Seattle, United States. Association for
  Computational Linguistics.

\bibitem[{Cao and Daum{\'e}~III(2020)}]{cao-daume-iii-2020-toward}
Yang~Trista Cao and Hal Daum{\'e}~III. 2020.
\newblock \href {https://doi.org/10.18653/v1/2020.acl-main.418} {Toward
  gender-inclusive coreference resolution}.
\newblock In \emph{Proceedings of the 58th Annual Meeting of the Association
  for Computational Linguistics}, pages 4568--4595, Online. Association for
  Computational Linguistics.

\bibitem[{Cho et~al.(2019)Cho, Kim, Kim, and Kim}]{cho-etal-2019measuring}
Won~Ik Cho, Ji~Won Kim, Seok~Min Kim, and Nam~Soo Kim. 2019.
\newblock \href {https://doi.org/10.18653/v1/W19-3824} {On {M}easuring {G}ender
  bias in {T}ranslation of {G}ender-neutral {P}ronouns}.
\newblock In \emph{Proceedings of the First Workshop on Gender Bias in Natural
  Language Processing}, pages 173--181, Florence, IT. Association for
  Computational Linguistics.

\bibitem[{Choubey et~al.(2021)Choubey, Currey, Mathur, and
  Dinu}]{choubey2021improving}
Prafulla~Kumar Choubey, Anna Currey, Prashant Mathur, and Georgiana Dinu. 2021.
\newblock Improving gender translation accuracy with filtered self-training.
\newblock \emph{arXiv preprint arXiv:2104.07695}.

\bibitem[{Costa-juss{\`a} and de~Jorge(2020)}]{costa-jussa-de-jorge-2020-fine}
Marta~R. Costa-juss{\`a} and Adri{\`a} de~Jorge. 2020.
\newblock \href {https://aclanthology.org/2020.gebnlp-1.3} {Fine-tuning neural
  machine translation on gender-balanced datasets}.
\newblock In \emph{Proceedings of the Second Workshop on Gender Bias in Natural
  Language Processing}, pages 26--34, Barcelona, Spain (Online). Association
  for Computational Linguistics.

\bibitem[{Dev et~al.(2021)Dev, Monajatipoor, Ovalle, Subramonian, Phillips, and
  Chang}]{dev-etal-2021-harms}
Sunipa Dev, Masoud Monajatipoor, Anaelia Ovalle, Arjun Subramonian, Jeff
  Phillips, and Kai-Wei Chang. 2021.
\newblock \href {https://doi.org/10.18653/v1/2021.emnlp-main.150} {Harms of
  gender exclusivity and challenges in non-binary representation in language
  technologies}.
\newblock In \emph{Proceedings of the 2021 Conference on Empirical Methods in
  Natural Language Processing}, pages 1968--1994, Online and Punta Cana,
  Dominican Republic. Association for Computational Linguistics.

\bibitem[{Gaido et~al.(2020)Gaido, Savoldi, Bentivogli, Negri, and
  Turchi}]{gaido-etal-2020-breeding}
Marco Gaido, Beatrice Savoldi, Luisa Bentivogli, Matteo Negri, and Marco
  Turchi. 2020.
\newblock \href {https://doi.org/10.18653/v1/2020.coling-main.350} {{Breeding
  Gender-aware Direct Speech Translation Systems}}.
\newblock In \emph{Proceedings of the 28th International Conference on
  Computational Linguistics}, pages 3951--3964, Online. International Committee
  on Computational Linguistics.

\bibitem[{Gaido et~al.(2021)Gaido, Savoldi, Bentivogli, Negri, and
  Turchi}]{gaido-etal-2021-split}
Marco Gaido, Beatrice Savoldi, Luisa Bentivogli, Matteo Negri, and Marco
  Turchi. 2021.
\newblock \href {https://doi.org/10.18653/v1/2021.findings-acl.313} {How to
  split: the effect of word segmentation on gender bias in speech translation}.
\newblock In \emph{Findings of the Association for Computational Linguistics:
  ACL-IJCNLP 2021}, pages 3576--3589, Online. Association for Computational
  Linguistics.

\bibitem[{Ghosh and Caliskan(2023)}]{ghosh2023chatgpt}
Sourojit Ghosh and Aylin Caliskan. 2023.
\newblock \href {http://arxiv.org/abs/2305.10510} {Chatgpt perpetuates gender
  bias in machine translation and ignores non-gendered pronouns: Findings
  across bengali and five other low-resource languages}.

\bibitem[{Ginel and Theroine(2022)}]{rivas-ginel-2022-all-inclusive}
María Isabel~Rivas Ginel and Sarah Theroine. 2022.
\newblock Neutralising for equality: All-inclusive games machine translation.
\newblock In \emph{Proceedings of New Trends in Translation and Technology},
  pages 125--133. NeTTT.

\bibitem[{H{\"o}glund and Flinkfeldt(2023)}]{hoglund2023gendering}
Frida H{\"o}glund and Marie Flinkfeldt. 2023.
\newblock De-gendering parents: Gender inclusion and standardised language in
  screen-level bureaucracy.
\newblock \emph{International Journal of Social Welfare}.

\bibitem[{Lardelli and Gromann(2022)}]{lardelli-gromann-2022-gender-fair}
Manuel Lardelli and Dagmar Gromann. 2022.
\newblock Gender-fair (machine) translation.
\newblock In \emph{Proceedings of New Trends in Translation and Technology},
  pages 166--177. NeTTT.

\bibitem[{Lauscher et~al.(2022)Lauscher, Crowley, and
  Hovy}]{lauscher-etal-2022-welcome}
Anne Lauscher, Archie Crowley, and Dirk Hovy. 2022.
\newblock \href {https://aclanthology.org/2022.coling-1.105} {Welcome to the
  modern world of pronouns: Identity-inclusive natural language processing
  beyond gender}.
\newblock In \emph{Proceedings of the 29th International Conference on
  Computational Linguistics}, pages 1221--1232, Gyeongju, Republic of Korea.
  International Committee on Computational Linguistics.

\bibitem[{Lauscher et~al.(2023)Lauscher, Nozza, Crowley, Miltersen, and
  Hovy}]{lauscher2023em}
Anne Lauscher, Debora Nozza, Archie Crowley, Ehm Miltersen, and Dirk Hovy.
  2023.
\newblock \href {http://arxiv.org/abs/2305.16051} {What about em? how
  commercial machine translation fails to handle (neo-)pronouns}.

\bibitem[{McConnell-Ginet(2013)}]{McConnell-Ginet2013}
Sally McConnell-Ginet. 2013.
\newblock \href {https://doi.org/10.1515/9783110307337.3} {Gender and its
  {R}elation to {S}ex: {T}he {M}yth of `{N}atural' {G}ender}.
\newblock In Greville~G. Corbett, editor, \emph{The Expression of Gender},
  pages 3--38. De Gruyter Mouton, Berlin, DE.

\bibitem[{Ovalle et~al.(2023)Ovalle, Goyal, Dhamala, Jaggers, Chang, Galstyan,
  Zemel, and Gupta}]{ovalle2023m}
Anaelia Ovalle, Palash Goyal, Jwala Dhamala, Zachary Jaggers, Kai-Wei Chang,
  Aram Galstyan, Richard Zemel, and Rahul Gupta. 2023.
\newblock “i’m fully who i am”: Towards centering transgender and
  non-binary voices to measure biases in open language generation.
\newblock In \emph{Proceedings of the 2023 ACM Conference on Fairness,
  Accountability, and Transparency}, pages 1246--1266.

\bibitem[{Piergentili et~al.(2023)Piergentili, Fucci, Savoldi, Bentivogli, and
  Negri}]{piergentili-etal-2023-gender}
Andrea Piergentili, Dennis Fucci, Beatrice Savoldi, Luisa Bentivogli, and
  Matteo Negri. 2023.
\newblock \href {https://aclanthology.org/2023.gitt-1.7} {Gender neutralization
  for an inclusive machine translation: from theoretical foundations to open
  challenges}.
\newblock In \emph{Proceedings of the First Workshop on Gender-Inclusive
  Translation Technologies}, pages 71--83, Tampere, Finland. European
  Association for Machine Translation.

\bibitem[{Rarrick et~al.(2023)Rarrick, Naik, Mathur, Poudel, and
  Chowdhary}]{rarrick2023gate}
Spencer Rarrick, Ranjita Naik, Varun Mathur, Sundar Poudel, and Vishal
  Chowdhary. 2023.
\newblock \href {http://arxiv.org/abs/2303.03975} {Gate: A challenge set for
  gender-ambiguous translation examples}.

\bibitem[{Renduchintala and
  Williams(2022)}]{renduchintala-williams-2022-investigating}
Adithya Renduchintala and Adina Williams. 2022.
\newblock \href {https://doi.org/10.18653/v1/2022.acl-long.243} {Investigating
  failures of automatic translationin the case of unambiguous gender}.
\newblock In \emph{Proceedings of the 60th Annual Meeting of the Association
  for Computational Linguistics (Volume 1: Long Papers)}, pages 3454--3469,
  Dublin, Ireland. Association for Computational Linguistics.

\bibitem[{Saunders et~al.(2020)Saunders, Sallis, and
  Byrne}]{saunders-etal-2020-neural}
Danielle Saunders, Rosie Sallis, and Bill Byrne. 2020.
\newblock \href {https://aclanthology.org/2020.gebnlp-1.4} {Neural machine
  translation doesn{'}t translate gender coreference right unless you make it}.
\newblock In \emph{Proceedings of the Second Workshop on Gender Bias in Natural
  Language Processing}, pages 35--43, Barcelona, Spain (Online). Association
  for Computational Linguistics.

\bibitem[{Saunders et~al.(2022)Saunders, Sallis, and
  Byrne}]{saunders-etal-2022-first}
Danielle Saunders, Rosie Sallis, and Bill Byrne. 2022.
\newblock \href {https://doi.org/10.18653/v1/2022.findings-acl.301} {First the
  worst: Finding better gender translations during beam search}.
\newblock In \emph{Findings of the Association for Computational Linguistics:
  ACL 2022}, pages 3814--3823, Dublin, Ireland. Association for Computational
  Linguistics.

\bibitem[{Savoldi et~al.(2021)Savoldi, Gaido, Bentivogli, Negri, and
  Turchi}]{savoldi2021gender}
Beatrice Savoldi, Marco Gaido, Luisa Bentivogli, Matteo Negri, and Marco
  Turchi. 2021.
\newblock \href {https://doi.org/10.1162/tacl_a_00401} {Gender bias in machine
  translation}.
\newblock \emph{Transactions of the Association for Computational Linguistics},
  9:845--874.

\bibitem[{Savoldi et~al.(2022{\natexlab{a}})Savoldi, Gaido, Bentivogli, Negri,
  and Turchi}]{savoldi-etal-2022-dynamics}
Beatrice Savoldi, Marco Gaido, Luisa Bentivogli, Matteo Negri, and Marco
  Turchi. 2022{\natexlab{a}}.
\newblock \href {https://doi.org/10.18653/v1/2022.gebnlp-1.12} {On the dynamics
  of gender learning in speech translation}.
\newblock In \emph{Proceedings of the 4th Workshop on Gender Bias in Natural
  Language Processing (GeBNLP)}, pages 94--111, Seattle, Washington.
  Association for Computational Linguistics.

\bibitem[{Savoldi et~al.(2022{\natexlab{b}})Savoldi, Gaido, Bentivogli, Negri,
  and Turchi}]{savoldi-etal-2022-morphosyntactic}
Beatrice Savoldi, Marco Gaido, Luisa Bentivogli, Matteo Negri, and Marco
  Turchi. 2022{\natexlab{b}}.
\newblock \href {https://doi.org/10.18653/v1/2022.acl-long.127} {Under the
  morphosyntactic lens: A multifaceted evaluation of gender bias in speech
  translation}.
\newblock In \emph{Proceedings of the 60th Annual Meeting of the Association
  for Computational Linguistics (Volume 1: Long Papers)}, pages 1807--1824,
  Dublin, Ireland. Association for Computational Linguistics.

\bibitem[{Silveira(1980)}]{silveira1980generic}
Jeanette Silveira. 1980.
\newblock \href
  {https://www.sciencedirect.com/science/article/pii/S0148068580921132}
  {Generic {M}asculine {W}ords and {T}hinking}.
\newblock \emph{Women's Studies International Quarterly}, 3(2-3):165--178.

\bibitem[{Stahlberg et~al.(2007)Stahlberg, Braun, Irmen, and
  Sczesny}]{stahlberg2007representation}
Dagmar Stahlberg, Friederike Braun, Lisa Irmen, and Sabine Sczesny. 2007.
\newblock \href {https://psycnet.apa.org/record/2007-01308-006}
  {{Representation of the Sexes in Language}}.
\newblock \emph{Social communication}, pages 163--187.

\bibitem[{Stanczak and Augenstein(2021)}]{stanczak2021survey}
Karolina Stanczak and Isabelle Augenstein. 2021.
\newblock {A Survey on Gender Bias in Natural Language Processing}.
\newblock \emph{arXiv preprint arXiv:2112.14168}.

\bibitem[{Stanovsky et~al.(2019)Stanovsky, Smith, and
  Zettlemoyer}]{stanovsky-etal-2019-evaluating}
Gabriel Stanovsky, Noah~A. Smith, and Luke Zettlemoyer. 2019.
\newblock \href {https://doi.org/10.18653/v1/P19-1164} {Evaluating gender bias
  in machine translation}.
\newblock In \emph{Proceedings of the 57th Annual Meeting of the Association
  for Computational Linguistics}, pages 1679--1684, Florence, Italy.
  Association for Computational Linguistics.

\bibitem[{Sun et~al.(2019)Sun, Gaut, Tang, Huang, ElSherief, Zhao, Mirza,
  Belding, Chang, and Wang}]{sun-etal-2019}
Tony Sun, Andrew Gaut, Shirlyn Tang, Yuxin Huang, Mai ElSherief, Jieyu Zhao,
  Diba Mirza, Elizabeth Belding, Kai-Wei Chang, and William~Yang Wang. 2019.
\newblock \href {https://doi.org/10.18653/v1/P19-1159} {Mitigating {G}ender
  {B}ias in {N}atural {L}anguage {P}rocessing: {L}iterature {R}eview}.
\newblock In \emph{Proceedings of the 57th Annual Meeting of the Association
  for Computational Linguistics}, pages 1630--1640, Florence, IT. Association
  for Computational Linguistics.

\bibitem[{Vanmassenhove et~al.(2021)Vanmassenhove, Emmery, and
  Shterionov}]{vanmassenhove-etal-2021-neutral}
Eva Vanmassenhove, Chris Emmery, and Dimitar Shterionov. 2021.
\newblock \href {https://doi.org/10.18653/v1/2021.emnlp-main.704} {{N}eu{T}ral
  {R}ewriter: {A} rule-based and neural approach to automatic rewriting into
  gender neutral alternatives}.
\newblock In \emph{Proceedings of the 2021 Conference on Empirical Methods in
  Natural Language Processing}, pages 8940--8948, Online and Punta Cana,
  Dominican Republic. Association for Computational Linguistics.

\end{thebibliography}
\bibliographystyle{acl_natbib}

\appendix


\end{document}